\documentclass[letterpaper, 10 pt, conference]{ieeeconf}  

\IEEEoverridecommandlockouts                              

\usepackage{kotex}
\usepackage{amsmath} 
\usepackage{amssymb}  
\usepackage{booktabs}
\usepackage{graphicx}
\usepackage{amsmath}
\usepackage{multirow}
\usepackage{xcolor}
\definecolor{bracecolor}{RGB}{200, 95, 35}
\usepackage{cuted}
\usepackage{capt-of}

\title{\LARGE \bf
\textcolor{bracecolor}{IORA}: 
\textcolor{bracecolor}{I}rregular 
\textcolor{bracecolor}{O}bject Immobilization under 
\textcolor{bracecolor}{R}estricted 
\textcolor{bracecolor}{A}ctuation
}

\author{Sang-Hun Han$^{1}$, Changyeon Won$^{1}$, JunGyu Lee$^{2}$, Min-Gyu Park$^{3,4}$, and Hae-Gon Jeon$^{2,\dagger}$
\thanks{\hspace*{-1em}$^{1}$Gwangju Institute of Science and Technology, $^{2}$Yonsei University,}%
\thanks{\hspace*{-1em}$^{3}$Korea Electronics Technology Institute, $^{4}$Polygom. $^{\dagger}$Corresponding author.}%
\thanks{\hspace*{-1em}{\tt\small sanghunhan@gm.gist.ac.kr}}%
}

\begin{document}

\maketitle

\thispagestyle{empty}
\pagestyle{empty}

\begin{strip}
  \centering
\vspace{-20mm}

  \includegraphics[width=\textwidth]{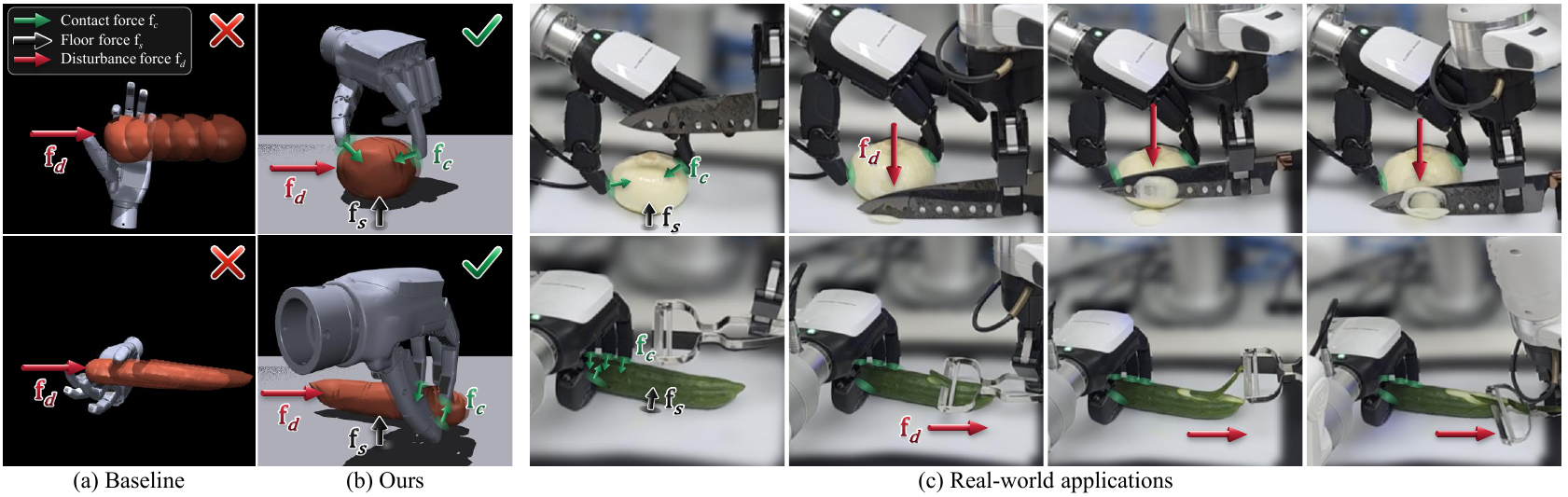} 
  \vspace{-6mm}
  \captionof{figure}{\textbf{Support-aware object immobilization.}
Under external disturbances, (a) baseline grasps fail to maintain the object pose, whereas (b) our grasps combine hand--object contact forces with tabletop reactions to stabilize it.
(c) The resulting stability enables real-world cutting and peeling.}
  \label{fig:teaser}
  \vspace{-4mm}
\end{strip}


\begin{abstract}

Stable grasping under task-specific disturbances is important for robotic manipulation. 
However, most dexterous grasp synthesis methods are developed for highly actuated hands in free space, and low-DoF hands may not be able to realize the contact topologies required by these grasps. 
Meanwhile, support surfaces can provide useful contact forces, but are often treated only as geometric constraints.
We present a support-aware grasp synthesis framework for object immobilization with low-DoF coupled hands. 
Given irregular object geometry, our method represents both hand and object local surface regions as graphs and searches for kinematically feasible matchings between them.
We then refine the wrist pose and motor state for geometric contact feasibility and validates the refined grasps through physical rollouts. 
Starting from the validated grasp pose, an optional task-guided local search explores nearby grasp configurations by varying the wrist pose and motor state, and selects one according to the specified task condition.
Experiments on diverse objects show that our method generates various grasp configurations that achieve high grasp success and strong disturbance resistance.
Real-world experiments further demonstrate physically executable grasps which enable object immobilization in cooking scenario.

\end{abstract}

\section{INTRODUCTION}
\label{sec:introduction}


Robotic grasping is a fundamental prerequisite for manipulation. 
Robotic hands are widely used to grasp and manipulate objects in various applications such as tool use, assembly, and household tasks.
Existing grasp methods~\cite{miller2004graspit,liu2022dfc, wang2023dexgraspnet} have primarily focused on automatically synthesizing grasp poses that enable the hand to successfully grasp and hold an object.
Especially, previous works~\cite{chen2025bodex, chen2025dexonomy} try to synthesize grasp to stable from disturbances for high-DoF dexterous hands, whose joints are independently actuated.

However, their practical applicability is limited by two factors. First, they require high-DoF hands, which are cost-prohibitive, making them impractical for widespread deployment. Second, they pay less attention to maintaining stability, resulting in vulnerable grasps under external forces. In real-world scenarios, such as holding an object while cutting, the stability of the object's pose is crucial for task success. 

While low-DoF hands offer a cost-effective alternative for widespread deployment, their coupled actuation introduces additional challenges for grasp synthesis. 
Due to the coupled joints, moving one finger region toward a desired contact may simultaneously lead to undesired movement in other regions. Furthermore, the planned joint motions may not be compatible due to the limited DoF. 
These undesired motions cause unintended contacts that may push or displace the object during grasping. 
To address these challenges, our key insight is to use the supporting surface as an additional source of contact for low-DoF grasping. 
Rather than relying on the hand alone to generate all contact forces, the object can be immobilized jointly by the hand and the support surface~\cite{hou2020sharedgrasping}. 
The surface provides passive normal and frictional forces without additional actuation~\cite{chavandafle2014extrinsic,eppner2015planning}, supplying opposition that is difficult for a coupled hand to generate by itself. 
With support from the surface, a low-DoF hand can achieve strong resistance to external forces even with simple grasp configurations.
We thus treat the support surface not only as a collision constraint, but as an active part of grasp synthesis and physical validation.
As illustrated in Fig.~\ref{fig:teaser}, the resulting support-aware grasps maintain the object pose under external disturbances and enable downstream tasks such as cutting and peeling.



%

\begin{figure*}[t]
    \centering
    \vspace{2mm}
    \includegraphics[clip=true, width=.98\linewidth]{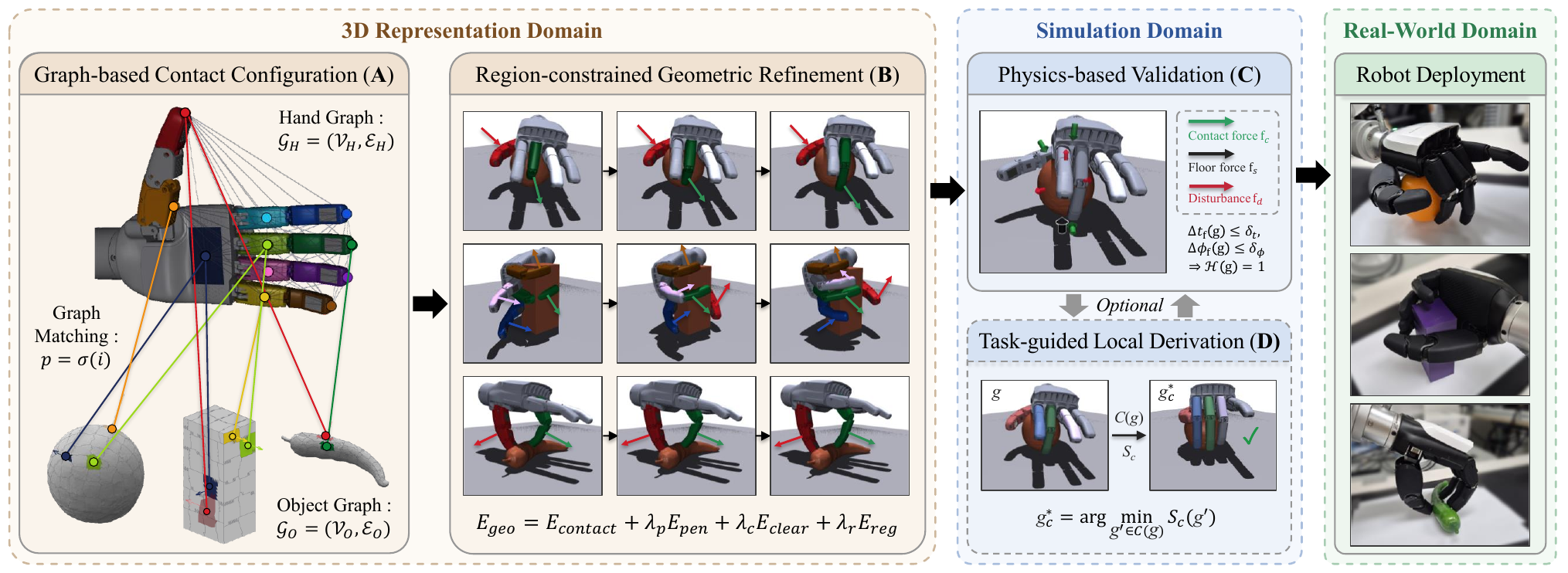}
    \vspace{-3mm}
    \caption{\textbf{Overview of our grasp synthesis framework.} Given the object geometry and a low-DoF coupled hand, we first search for kinematically feasible correspondences between hand and object regions using graph-based contact matching. The resulting grasp candidates are geometrically refined while preserving their contact correspondences, and physically validated under external disturbances with support from the tabletop. Starting from a validated grasp, an optional task-guided local search further adapts the wrist pose and motor state to task-specific requirements.} 
    \vspace{-6mm}
    \label{fig:overview}
\end{figure*}

As shown in Fig.~\ref{fig:overview}, our grasp synthesis pipeline comprises three stages followed by an optional task-guided local derivation.
First, given a robotic hand embodiment and object geometry, we represent hand regions and local object surface regions as graphs, based on distance and surface-normal relations between each regions. We then search for multiple feasible correspondences between the two graphs to generate diverse grasp candidates, determining which hand region should contact each object region geometry.
%
After that, we locally refine the resulting grasp initialization while preserving their correspondences, minimizing penetration and separation between the matched contact regions. 
Finally, we evaluates each refined grasp under external disturbances in physics and retains only the grasps that pass the disturbance tests.
Optionally, starting from a validated grasp, the local search derives nearby configurations according to deployment-specific task conditions without repeating the global contact search.

We evaluate the proposed framework on objects with irregular geometries and scale under directional disturbances. 
We compare the synthesized grasps with existing dexterous grasp synthesis methods and evaluate sustainable disturbance resistance and contact quality. 
We further validate our grasps on the RH56 hand robot with real objects, demonstrating diverse grasp configurations and robustness to external disturbances. We also demonstrate that these grasps can support downstream applications such as cooking by stably immobilizing objects during cutting and slicing.

\section{RELATED WORK}
\label{sec:related_work}

\subsection{Dexterous Grasp Synthesis}

Classical dexterous grasp synthesis searches for stable hand configurations using analytical measures such as force closure and the Ferrari-Canny metric~\cite{ferrari1992planning,miller2004graspit}. Differentiable formulations improve the scalability and physical fidelity of this search~\cite{liu2022dfc,wang2023dexgraspnet,chen2024springgrasp}, while recent QP-based methods optimize contact-force feasibility~\cite{chen2025bodex,zurbrugg2025graspqp} and Dexonomy~\cite{chen2025dexonomy} broadens the range of synthesized grasp types. However, these methods treat contact forces as free variables bounded only by friction cones. Simulation is used primarily to filter candidates or train grasp evaluators~\cite{wang2023dexgraspnet,chen2025bodex,chen2025dexonomy,lum2024get}, and, apart from the fingertip compliance modeled by SpringGrasp~\cite{chen2024springgrasp}, actuator-to-contact-force transmission remains unmodeled, so a certified force distribution need not be realizable by the hand. Task-oriented methods additionally evaluate grasps against downstream wrench requirements~\cite{li1988taskoriented,borst2004taskwrench}. For example, TaskDexGrasp~\cite{chen2024taskdexgrasp} optimizes the relation between a prescribed Task Wrench Space (TWS) and the Grasp Wrench Space (GWS).

Hierarchical Fingertip Space~\cite{hang2016hfts} organizes the grasp search space by discretizing object contact regions, and Matak and Hermans~\cite{matak2023visualtactile} restrict contacts to tactile-sensing regions. Cross-embodiment methods transfer contact intent from human demonstrations~\cite{lakshmipathy2022contacttransfer,qi2026genhand} or infer hand-link--object-region correspondences from learned priors~\cite{attarian2023geomatch,wei2025drograsp,fei2025trograsp}. Although subsequent optimization may improve feasibility on the target hand, the contact structure itself is not derived from its coupled actuation. In contrast, we generate contact topologies directly from the geometry and coupled actuation of the target hand, determining executability during synthesis without relying on a learned prior or seed grasp.

\vspace{-1mm}
\subsection{Underactuated and Coupled Hands}
\label{sec:underactuated}

Reducing the number of independent actuators has long been studied through postural synergies, mechanical transmissions, and underactuated hand designs. Ciocarlie and Allen~\cite{ciocarlie2009subspaces} optimize grasp configurations in low-dimensional posture subspaces. Prattichizzo et al.~\cite{prattichizzo2010synergies} analyze how postural synergies affect motion and force controllability, and Gabiccini and Bicchi~\cite{gabiccini2010synergies} show that the available synergies also constrain the optimal choice of grasping forces. Together, these results indicate that reduced actuation restricts reachable postures and realizable contact forces at the same time. 
Birglen and Gosselin~\cite{birglen2004kinetostatic} analyze how a fixed transmission determines the contact forces an underactuated finger can produce along its phalanges, and Pozzi et al.~\cite{pozzi2017quality} introduce grasp-quality measures that account for the force directions controllable by underactuated and compliant hands.
These formulations treat coupling as a constraint on the forces achievable for a given contact set, whereas we make it a generation-time condition that retains only contact sets admitting a common coupled motor configuration.

\vspace{-1mm}
\subsection{Environment-Assisted Grasping}

Robots can exploit contacts with the surrounding environment to extend the capabilities of the hand itself. Extrinsic Dexterity~\cite{chavandafle2014extrinsic} uses gravity and external surfaces for manipulation with simple grippers. Kazemi et al.~\cite{kazemi2012support} show that deliberately interacting with a supporting table can be critical for robustly acquiring small objects, while Eppner et al.~\cite{eppner2015planning} develop grasp strategies that exploit environmental constraints rather than avoiding them. Manipulation with Shared Grasping~\cite{hou2020sharedgrasping} goes further by defining a shared grasp in which contacts from the hand and environment jointly support the object, allowing environmental contacts to substitute for hand contacts.

Environment contacts have also been integrated into contact-rich planning, where pre-grasp motions or sliding and pivoting trajectories are optimized together with the supporting forces they rely on~\cite{chen2023nonprehensile,zhang2023stocs}. BODex~\cite{chen2025bodex} includes a tabletop grasp dataset and table-aware collision-free trajectories, but its grasp-force optimization is formulated over hand-object contact wrenches and treats the table as a collision constraint rather than a load-bearing contact.
These works use the environment to acquire or move an object, whereas we synthesize a static grasp in which the supporting surface remains a load-bearing contact that resists external disturbances together with the hand during both synthesis and physical validation.

\section{Method}
\label{sec:method}


We propose a grasp synthesis framework for a low-DoF hand that maintain an object's pose on a supporting surface like tabletop under external disturbances. 

A grasp motion is represented by $\mathbf{g}=(\mathbf{T},\mathbf{u})$, where $\mathbf{T}\in SE(3)$ is the wrist pose and $\mathbf{u}\in\mathbb{R}^{m}$ denotes the states of the $m$ actuated motors.
The corresponding joint angles $\mathbf{q}\in\mathbb{R}^{n}$ are determined by the fixed coupling:
\begin{equation}
\mathbf{q}=\mathbf{C}\mathbf{u},
\qquad
\mathbf{C}\in\mathbb{R}^{n\times m},
\label{eq:coupled_kinematics}
\end{equation}
where $n$ denote the numbers of joints and $\mathbf{C}$ is a linear mapping from $\mathbf{u}$ to $\mathbf{q}$. 

%
%

\vspace{-1mm}
\subsection{Graph-Based Contact Configuration}
\label{sec:association}
Instead of relying on data-driven grasp priors, we synthesize diverse grasp candidates by explicitly exploring different hand-object contact configurations. 
Specifically, we represent the hand contact regions and object surface regions as graphs, denoted by $\mathcal{G}=(\mathcal{V},\mathcal{E})$, where $\mathcal{V}$ and $\mathcal{E}$ denote the nodes and edges, respectively. 
We then search for feasible matchings between the two graphs, where each matching defines a different contact configuration for grasp initialization.
We represent the hand and object graphs as
\begin{equation}
\mathcal{G}_{H}=(\mathcal{V}_{H},\mathcal{E}_{H}),
\qquad
\mathcal{G}_{O}=(\mathcal{V}_{O},\mathcal{E}_{O}),
\label{eq:graph_representation}
\end{equation} 
where $\mathcal{V}_{H}$ contains contact-capable hand regions, including fingertip, finger proximal, and palm surfaces, while $\mathcal{V}_{O}$ contains local surface regions of the object. 
The object surface is partitioned based on spatial distance and normal consistency, with each region is considered a candidate for a contact region. 
The edges $\mathcal{E}_{H}$ and $\mathcal{E}_{O}$ encode the geometric relations between regions using their distance $d$ and relative surface-normal angle $\theta$. 


A contact configuration is feasible only when the geometric relations of the selected hand regions match those of the assigned object regions. We therefore compare their pairwise distance and relative surface-normal angle, and retain only configurations realizable by a common motor state $\mathbf{u}$. This provides a direct geometric and kinematic feasibility under the hand coupling.
Let $\mathcal{U}$ denote the sampled motor states of the coupled hand, and $\mathbf{r}=(d,\theta)$ denote each pairwise geometric relation.
For a pair of hand regions $(i,j)$ and object surface regions $(p,q)$, we define
\begin{equation}
\begin{aligned}
\mathbf{r}^{H}_{ij}(\mathbf{u})
&= \big(d^{H}_{ij}(\mathbf{u}),\theta^{H}_{ij}(\mathbf{u})\big),
\qquad \mathbf{u}\in\mathcal{U},\\
\mathbf{r}^{O}_{pq}
&= \big(\mathcal{D}^{O}_{pq},\theta^{O}_{pq}\big),
\qquad \mathcal{D}^{O}_{pq}=\big[\,d^{\min}_{pq},\,d^{\max}_{pq}\,\big],
\end{aligned}
\label{eq:pair_relations}
\end{equation}
where $d^{H}_{ij}(\mathbf{u})$ and $\theta^{H}_{ij}(\mathbf{u})$ denote the distance and normal angle between the two hand regions under motor state $\mathbf{u}$, while $\mathcal{D}^{O}_{pq}$ and $\theta^{O}_{pq}$ denote the distance range and normal angle between the two object regions. 
Note that $\mathcal{D}^{O}_{pq}$ is a distance range because the contact locations can vary within the two object regions.

We search for multiple matchings $\sigma$ between hand and object regions to generate diverse grasp candidates, where $p=\sigma(i)$ denotes the object region $p$ assigned to hand region $i$.
A hand-region pair $(i,j)$ is compatible with its matched object-region pair when their pairwise distance and relative normal angle satisfy predefined tolerances. For a coupled hand, however, pairwise compatibility alone is insufficient because different pairs may require different motor states. 
We therefore require a single motor state that satisfies all matched pairwise relations:
\begin{equation}
\begin{aligned}
& d^{H}_{ij}(\mathbf{u}) \in \mathcal{D}^{O}_{\sigma(i)\sigma(j)}(\tau_d),\\
& \big|\theta^{H}_{ij}(\mathbf{u})-\theta^{O}_{\sigma(i)\sigma(j)}\big| \le \tau_{\theta},
\qquad \forall\, i<j,
\end{aligned}
\label{eq:common_witness}
\end{equation}
for some $\mathbf{u}\in\mathcal{U}$ shared by all pairs.

Here, $\tau_d$ and $\tau_{\theta}$ are fixed tolerances that account for the sampling of $\mathcal{U}$ and the within-region variation of contact points and normals, with $\mathcal{D}^{O}_{pq}(\tau_d)=\big[\,d^{\min}_{pq}-\tau_d,\;d^{\max}_{pq}+\tau_d\,\big]$.

We explore different hand-region subsets and expand their object-region matchings using the pairwise compatibility above, while retaining multiple feasible solutions to preserve grasp diversity. 
Each feasible matching is converted into an initial grasp by selecting a compatible motor state and aligning the matched hand and object regions.


\vspace{-1mm}
\subsection{Region-Constrained Geometric Refinement}
\label{sec:geometric_refinement}

The graph-based initialization is kinematically consistent but may still contain penetration or separation between the matched contact regions. 
We therefore locally refine the grasp $\mathbf{g}=(\mathbf{T},\mathbf{u})$ while keeping the correspondence $\sigma$ fixed:
\begin{equation}
\begin{aligned}
\mathbf{g}^{*}
&=
\arg\min_{\mathbf{g}}\; E_{\mathrm{geo}}(\mathbf{g};\sigma),\\
E_{\mathrm{geo}}
&=
E_{\mathrm{contact}}
+\lambda_{p}E_{\mathrm{pen}}
+\lambda_{c}E_{\mathrm{clear}}
+\lambda_{r}E_{\mathrm{reg}} .
\end{aligned}
\label{eq:geometric_refinement}
\end{equation}
$E_{\mathrm{contact}}$ reduces the separation between matched hand and object regions while encouraging their surfaces to face each other. $E_{\mathrm{pen}}$ penalizes penetration into the object and support surface, while $E_{\mathrm{clear}}$ suppresses unintended contact from non-matched hand regions. $E_{\mathrm{reg}}$ limits changes in the wrist pose and motor state from the initialization, keeping the refinement local. Since the correspondence $\sigma$ is fixed and the joint configuration follows the coupled kinematics in Eq.~\eqref{eq:coupled_kinematics}, the refinement improves geometric contact feasibility without leaving the realizable configuration space of the hand.

The correspondence $\sigma$ remains unchanged during refinement. The exact contact location may move within the matched object region, but the contact region assignment is preserved. Refined grasps are then filtered using geometric checks for contact orientation, object penetration, tabletop clearance, and task constraints before physical evaluation.

\begin{figure*}[ht]
  \centering
  \vspace{2mm}  \includegraphics[width=0.95\textwidth]{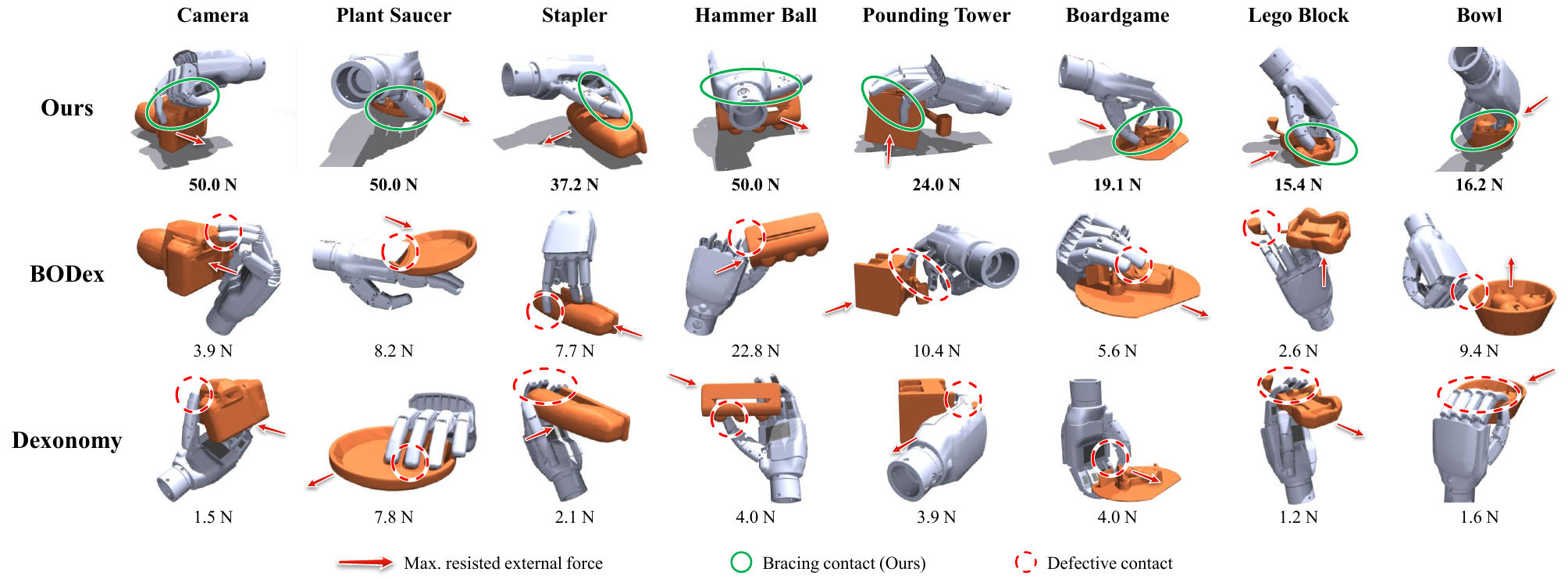}
 \vspace{-3mm}
  \caption{Qualitative comparison with BODex and Dexonomy on eight objects.
  For each grasp we apply external forces along the tested axis directions
  in MuJoCo and report the maximum force resisted along the weakest
  direction, capped at 50\,N. Red arrows show this weakest direction for
  each grasp. Green solid circles mark bracing contacts where our grasps
  wrap the object with multiple links and press it against the support
  surface. Red dashed circles mark defective contacts of the baselines
  such as fingertip-only touch, flat fingers resting on the surface, and
  one-sided contact without opposition.}
  \label{fig:compare}
  \vspace{-4mm}
\end{figure*}

\vspace{-1mm}
\subsection{Physics-Based Validation and Selection}
\label{sec:physics_validation}




Geometric validity alone does not guarantee that a grasp remains stable after holding.
We evaluate each refined candidate by replaying the grasp execution in simulation. The hand starts from a pre-grasp configuration near the planned contacts and closes toward the target grasp without moving the object.
During holding, the hand contact forces $\mathbf{f}_c$ and the support reaction $\mathbf{f}_s$ from the tabletop balance the object, while a constant disturbance $\mathbf{f}_d$ is applied along each direction in $\mathcal{F}_{6}=\{\pm\hat{\mathbf{x}},\pm\hat{\mathbf{y}},\pm\hat{\mathbf{z}}\}$ to evaluate grasp stability.


For a grasp $\mathbf{g}$ and the disturbance $\mathbf{f}_d\in\mathcal{F}_{6}$, let
$\Delta t_{\mathbf{f}}(\mathbf{g})$ and
$\Delta\phi_{\mathbf{f}}(\mathbf{g})$
denote the maximum object translation and rotation during the rollout.
We define a successful hold $\mathcal{H}$:
\begin{equation}
\begin{aligned}
\mathcal{H}(\mathbf{g})=1
\Longleftarrow\quad
&\Delta t_{\mathbf{f}}(\mathbf{g})\leq\delta_{t},\\
&\Delta\phi_{\mathbf{f}}(\mathbf{g})\leq\delta_{\phi},
\quad \forall \mathbf{f}\in\mathcal{F}_{6}.
\end{aligned}
\label{eq:physical_hold}
\end{equation}
where $\delta_{t}$ and $\delta_{\phi}$ are the allowed translation and rotation thresholds.
Candidates passing this test are rechecked for hand-object penetration and support-surface consistency, and retained across contact structures to preserve grasp diversity.



\vspace{-1mm}
\subsection{Optional Task-Guided Local Derivation}
\label{sec:derivation}
The preceding process yields a complete grasp without task-specific requirements. However, some tasks need the grasp to account for external factors beyond grasp stability. For example, when using a knife to cut an object, the hand should remain clear of the intended cutting region. Similarly, a grasp may need to better resist disturbances from a particular direction. In this process, we search grasps to accommodate such task-specific requirements.


We start from the obtained physically validated grasp through the preceding process, which provides a reliable starting point for task-specific derivation.
Given the validated grasp $\mathbf{g}=(\mathbf{T},\mathbf{u})$, we explore task-specific grasp candidates in its local neighborhood by applying perturbations to the wrist pose $\mathbf{T}$ and motor state $\mathbf{u}$, including small translations and discrete yaw rotations of the wrist, local scans of the motor state, and contact configuration.


%
We define $\mathcal{C}(\mathbf{g})$ as the set of locally perturbed grasps that satisfy the hard conditions, while $c$ specifies the task-specific soft conditions for ranking the feasible candidates.
The final variant is selected from this candidate set by
\begin{equation}
    \mathbf{g}^{*}_{c}
    =
    \arg\min_{\mathbf{g}'\in\mathcal{C}(\mathbf{g})}
    S_{c}(\mathbf{g}'),
\label{eq:task_derivation}
\end{equation}
where $S_{c}$ is a task-dependent ranking criterion for soft conditions $c$. 
The conditions $c$ specifies task preferences such as resistance to a particular disturbance direction, desired contact placement, or a preferred grasp style. Hard task constraints, such as keeping the hand outside a specified region, maintaining an unobstructed load path, or requiring additional fingers to contact the object, are enforced when constructing $\mathcal{C}_{c}(\mathbf{g})$. Among the resulting feasible candidates, $S_{c}$ ranks each grasp according to the specified task preference.
After that, we check geometric feasibility of the final grasp as described in Sec.~\ref{sec:physics_validation}.
This optional stage adapts an existing successful grasp toward application requirements while reusing the validated grasp as the starting point.

\begin{table*}[t]
\vspace{1mm}  
\caption{Quantitative comparison on the Low-DoF RH56 under floating and tabletop settings.}
\vspace{-3mm}
\centering
\renewcommand{\arraystretch}{1.18}
\scriptsize
\setlength{\tabcolsep}{2.35pt}
\resizebox{0.9\textwidth}{!}
{%
\begin{tabular}{ll|cc|ccccccc|cccc}
\hline
\noalign{\hrule height 1pt}
\multirow{2}{*}{\textbf{Setting}}
& \multirow{2}{*}{\textbf{Method}}
& \multicolumn{2}{c|}{\textbf{Success}}
& \multicolumn{7}{c|}{\textbf{Resistance}}
& \multicolumn{4}{c}{\textbf{Contact Quality}}
\\
\cline{3-15}
& & \textbf{GSR$\uparrow$}
& \textbf{OSR$\uparrow$}
& \textbf{$F_{\min}\uparrow$}
& \textbf{R@5}
& \textbf{R@10}
& \textbf{R@20}
& \textbf{R@30}
& \textbf{R@40}
& \textbf{R@50}
& \textbf{PD$\downarrow$}
& \textbf{SPD$\downarrow$}
& \textbf{CLN$\uparrow$}
& \textbf{CDC$\downarrow$}
\\
& & (\%) & (\%) & (N)
& (\%) & (\%) & (\%) & (\%) & (\%) & (\%)
& (mm, avg./med.) & (mm) & & (mm, med.) 
\\
\noalign{\hrule height 1pt}

\multirow{4}{*}{Tabletop}
& \textbf{Ours}
& 77.7 & \textbf{88.5}
& \textbf{10.9}
& \textbf{68.1} & \textbf{53.9} & \textbf{37.2}
& \textbf{27.5} & \textbf{22.5} & \textbf{18.6}
& 1.01 / 0.09 & 0.00 & \textbf{3.03} & 0.28
\\

& Dexonomy
& \textbf{80.2} & 42.5
& 3.6
& 37.4 & 18.2 & 2.7
& 1.6 & 0.5 & 0.5
& \textbf{0.04 / 0.01} & 0.00 & 3.02 & \textbf{0.25}
\\

& DexGraspNet
& 64.9 & 54.5
& 3.8
& 46.2 & 37.3 & 28.3
& 19.0 & 11.5 & 8.6
& 2.43 / 1.51 & 0.00 & 2.90 & 1.93
\\

& BODex
& 62.0 & 77.5
& 3.1
& 40.5 & 28.8 & 14.0
& 7.5 & 6.5 & 5.2
& 1.40 / 0.08 & 0.00 & 2.31 & 0.32
\\

\hline

\multirow{4}{*}{Floating}
& \textbf{Ours}
& \textbf{54.7} & \textbf{60.0}
& \textbf{2.3}
& \textbf{39.0} & \textbf{24.1} & \textbf{9.7}
& \textbf{4.5} & 1.0 & 0.8
& 0.10 / 0.09 & 0.00 & 3.14 & \textbf{0.01}
\\

& Dexonomy
& 49.7 & 28.5
& 1.7
& 19.3 & 6.4 & 0.5
& 0.5 & 0.0 & 0.0
& \textbf{0.09 / 0.08} & 0.00 & \textbf{3.80} & 0.01
\\

& DexGraspNet
& 31.9 & 30.0
& 0.3
& 18.6 & 9.7 & 4.7
& 3.2 & 1.4 & 1.1
& 5.77 / 6.80 & 0.00 & 3.42 & 6.81
\\

& BODex
& 16.2 & 26.5
& 0.3
& 9.8 & 7.0 & 3.8
& 2.5 & \textbf{2.5} & \textbf{2.0}
& 9.34 / 8.49 & 0.00 & 2.91 & 4.41
\\

\noalign{\hrule height 1pt}
\hline
\end{tabular}}
\vspace{-2mm}
\label{tab:rh56_main}
\end{table*}

\begin{table*}[t]
\vspace{-1mm}
\caption{Quantitative comparison on the High-DoF Shadow Hand under the no-selection protocol. 
}
\vspace{-3mm}

\centering
\renewcommand{\arraystretch}{1.18}
\scriptsize
\setlength{\tabcolsep}{2.4pt}
\resizebox{0.9\textwidth}{!}
{%
\begin{tabular}{ll|cc|ccccccc|cccc}
\hline
\noalign{\hrule height 1pt}
\multirow{2}{*}{\textbf{Scene}}
& \multirow{2}{*}{\textbf{Method}}
& \multicolumn{2}{c|}{\textbf{Success}}
& \multicolumn{7}{c|}{\textbf{Resistance}}
& \multicolumn{4}{c}{\textbf{Contact Quality}}
\\
\cline{3-15}
& & \textbf{GSR$\uparrow$}
& \textbf{OSR$\uparrow$}
& \textbf{$F_{\min}\uparrow$}
& \textbf{R@5}
& \textbf{R@10}
& \textbf{R@20}
& \textbf{R@30}
& \textbf{R@40}
& \textbf{R@50}
& \textbf{PD$\downarrow$}
& \textbf{SPD$\downarrow$}
& \textbf{CLN$\uparrow$}
& \textbf{CDC$\downarrow$}
\\
& & (\%) & (\%) & (N)
& (\%) & (\%) & (\%) & (\%) & (\%) & (\%)
& (mm, avg./med.) & (mm) & & (mm, med.) 
\\
\noalign{\hrule height 1pt}

\multirow{4}{*}{Tabletop}
& \textbf{Ours}
& \textbf{88.1} & 93.5
& \textbf{9.8}
& \textbf{64.1} & \textbf{49.6} & \textbf{34.0}
& \textbf{22.2} & \textbf{16.6} & \textbf{11.9}
& 0.83 / 0.41 & 0.06 & 3.96 & 0.07
\\

& Dexonomy
& 86.0 & 95.0
& 5.7
& 54.8 & 24.0 & 7.2 & 1.5 & 0.2 & 0.2
& 1.18 / 0.87 & \textbf{0.02} & 5.12 & \textbf{0.02}
\\

& DexGraspNet
& 81.5 & \textbf{95.5}
& 4.2
& 40.2 & 10.8 & 1.8 & 1.0 & 0.8 & 0.5
& 0.60 / 0.63 & 0.03 & \textbf{5.37} & \textbf{0.02}
\\

& BODex
& 50.2 & 73.5
& 1.6
& 17.2 & 2.2 & 0.2 & 0.0 & 0.0 & 0.0
& \textbf{0.19 / 0.08} & 0.03 & 4.57 & \textbf{0.02}
\\

\hline

\multirow{4}{*}{Floating}
& \textbf{Ours}
& \textbf{80.4} & 86.0
& \textbf{10.5}
& \textbf{64.0} & \textbf{51.8} & \textbf{33.2}
& \textbf{18.8} & \textbf{13.4} & \textbf{8.7}
& 1.13 / 0.94 & 0.10 & 5.21 & 0.23
\\

& Dexonomy
& 69.2 & \textbf{87.0}
& 3.5
& 36.5 & 16.0 & 3.5 & 0.2 & 0.0 & 0.0
& 1.15 / 0.89 & \textbf{0.02} & \textbf{5.50} & \textbf{0.02}
\\

& DexGraspNet
& 46.0 & 68.0
& 1.2
& 5.8 & 0.8 & 0.0 & 0.0 & 0.0 & 0.0
& 0.53 / 0.47 & 0.03 & 4.73 & \textbf{0.02}
\\

& BODex
& 19.0 & 29.5
& 0.5
& 4.0 & 0.2 & 0.0 & 0.0 & 0.0 & 0.0
& \textbf{0.27 / 0.09} & 0.03 & 4.83 & \textbf{0.02}
\\

\noalign{\hrule height 1pt}
\hline
\end{tabular}}
\vspace{-6mm}
\label{tab:shadow_nosel}
\end{table*}

\section{Experimental Results} 
\label{sec:experiments}

We extensively evaluate the proposed method in both simulation and real-world settings to assess grasp success ratio, physical robustness and generalization across diverse objects. We further provide controlled analyses to validate the contributions of support-aware synthesis and the individual components of our framework.


\vspace{-1mm}
\subsection{Experimental Setup}
\label{sec:exp_setup}

\vspace{-1mm}
\paragraph{Simulation setup}
We evaluate the proposed method on the coupled Inspire RH56 and the Shadow Hand using 200 randomly sampled objects from the DexGraspNet~\cite{wang2023dexgraspnet} test split. All methods use the same object set and are evaluated under a shared MuJoCo protocol. We use a $1$ms simulation step, $0.1$kg object mass, and a friction coefficient of $0.6$. 
The contact-configuration tolerances are fixed across all objects to $\tau_d=8$\,mm and $\tau_\theta=30^\circ$ for pairwise pruning in Eq.~\eqref{eq:common_witness}.
After reaching the grasp configuration,
we apply an additional finger closing command controlled by a squeeze coefficient, with $0.5$ used as the default setting.

\paragraph{Evaluation protocol}
We assess grasp quality using success, contact, and penetration metrics. Grasp Success Rate (GSR) is the percentage of attempted grasps that resist the prescribed disturbance, whereas Object Success Rate (OSR) is the percentage of objects for which at least one successful grasp is found. Penetration Depth (PD) and Self-Penetration Depth (SPD) denote the maximum hand--object and inter-link penetration depths, respectively. Contact Link Number (CLN) counts hand links within $2\,\mathrm{mm}$ of the object surface, while Contact Distance Consistency (CDC) is the range of signed contact distances across fingers, with lower values indicating more consistent contact.

For GSR and OSR, we apply constant forces along $\mathcal{E}=\{\pm\hat{\mathbf{x}},\pm\hat{\mathbf{y}},+\hat{\mathbf{z}}\}$.
A grasp succeeds if it resists a $0.98\,\mathrm{N}$ force for $2\,\mathrm{s}$ while maintaining translational and rotational displacements below $50\,\mathrm{mm}$ and $15^{\circ}$. We then increase the force to $50\,\mathrm{N}$ at $8\,\mathrm{N/s}$ under tighter bounds of $5\,\mathrm{mm}$ and $10^{\circ}$ and report
\begin{equation}
    F_{\min}(\mathbf{g})
    =
    \min_{\hat{\mathbf{e}}\in\mathcal{E}}
    F_{\max}^{\hat{\mathbf{e}}}(\mathbf{g}),
    \qquad
    R@F
    =
    \Pr\!\left(F_{\min}(\mathbf{g}) \geq F\right).
    \label{eq:force_resistance_metric}
\end{equation}
Here, $F_{\max}^{\hat{\mathbf{e}}}(\mathbf{g})$ is the maximum force resisted by grasp $\mathbf{g}$ along $\hat{\mathbf{e}}$. We compare against Dexonomy~\cite{chen2025dexonomy}, DexGraspNet~\cite{wang2023dexgraspnet}, and BODex~\cite{chen2025bodex} using their open-source implementations under the same MuJoCo evaluation protocol.

\vspace{-1mm}
\subsection{Comparison with Existing Methods}

\paragraph{Coupled RH56}
Fig.~\ref{fig:compare} demonstrates that our grasps form topologies that robustly hold complex objects in the DexGraspNet~\cite{wang2023dexgraspnet} dataset. In contrast, prior methods struggle to form topologies that can apply balanced forces.
Tab.~\ref{tab:rh56_main} quantitatively compares grasp synthesis methods with and without tabletop support. 
Our method typically shows the best performance in the tabletop and floating setting, with grasp success and disturbance resistance across the full force range. 
This is because our contact point search based on distance and angle contributes to stable object support.
Moreover, across all methods, the stronger resistance profile provided by the support surface shows that its benefit extends beyond grasp success and remains effective under increasingly large external disturbances.
Dexonomy shows a high GSR because difficult objects for which it fails to generate valid grasps are excluded from the GSR calculation, but these objects are still counted as failures in OSR, resulting in a lower OSR.
%
For PD, our grasps show comparable results to Dexonomy at matched grip forces (\(<20\,\mathrm{N}\)), as reflected by the similar median values.
The higher mean PD mainly comes from high-force cases above \(20\,\mathrm{N}\), where Dexonomy often fails and is therefore excluded from the PD calculation, whereas our grasps remain successful and are recorded with higher PD. Nevertheless, the median PD remains similar, indicating comparable penetration for typical successful grasps.

\paragraph{Generality across hand embodiments}
We further evaluate our method on the Shadow Hand using the same synthesis procedure. 
We conduct this experiment without the physics-based validation described in Sec.~\ref{sec:physics_validation}. Because, with this validation, both GSR and OSR exceed \(99\%\) for nearly all methods including ours.
As shown in Tab.~\ref{tab:shadow_nosel}, our method maintains high grasp success and strong disturbance resistance in both floating and tabletop settings, demonstrating that the method is not specific to the RH56 embodiment. 


\vspace{-2mm}
\subsection{Analysis and Ablations}
\label{sec:analysis_ablation}

\begin{table}[t]
  \vspace{2mm}

\caption{Analysis of the grasp synthesis pipeline on the coupled RH56 in the tabletop setting.}
  \vspace{-2mm}
\centering
\renewcommand{\arraystretch}{1.16}
\scriptsize
\setlength{\tabcolsep}{2.5pt}
\resizebox{0.9\linewidth}{!}
{%
\begin{tabular}{l|ccccccc}
\hline
\noalign{\hrule height 1pt}
\textbf{Variant}
& \textbf{GSR}
& \textbf{OSR}
& \textbf{$F_{\min}$}
& \textbf{R@5/10/20}
& \textbf{PD avg./med.}
& \textbf{CLN} \\
& & (\%) & (\%) & (N) & (\%) & (mm) & \\
\noalign{\hrule height 1pt}

Graph-based configuration
& 77.8 & 93.0 & 16.0
& 70.2/61.2/45.0
& 4.07/2.71 & 4.52 \\

+ Geometric refinement
& 60.5 & 79.5 & 5.3
& 50.2/40.8/28.2
& 3.15/0.16 & 3.24 \\

+ Physics validation
& \textbf{86.3} & \textbf{93.5} & \textbf{11.0}
& \textbf{71.5/53.1/31.9}
& \textbf{1.01/0.09} & \textbf{3.07} \\

\noalign{\hrule height 1pt}
\hline
\end{tabular}}
   \vspace{-3mm}
\label{tab:pipeline_ablation}
\end{table}

\paragraph{Analysis of the proposed pipeline}
Tab.~\ref{tab:pipeline_ablation} examines the role of each component in our pipeline.
The process in Sec.~\ref{sec:association} establishes geometrically stable contact configurations, leading to high grasp success, but often places the hand deeply around the object and causes substantial penetration resulting high PD. 
The process in Sec.~\ref{sec:geometric_refinement} preserves the contact matching while moving penetrated hand parts toward the object surface, thereby improving the most of synthesized grasps. 
However, without physical validation, penetration-free grasps may fail due to insufficient contact force, while deeply penetrated grasps can remain as successful cases. 
The process in Sec.~\ref{sec:physics_validation} addresses this issue by restoring grasp strength and filtering physically invalid grasps, resulting in robust grasps with minimal penetration resulting low PD. In summary, we first refines the most of synthesized grasps, and then filters the final successful grasps.



\begin{table}[t]
\caption{Static SOCP force-balance analysis of settled grasps with tabletop contacts included or removed.}
\vspace{-2mm}
\centering
\renewcommand{\arraystretch}{1.14}
\scriptsize
\setlength{\tabcolsep}{5.0pt}
\resizebox{0.90\linewidth}{!}
{%
\begin{tabular}{l|cc}
\hline
\noalign{\hrule height 1pt}
\textbf{Metric}
& \textbf{With support}
& \textbf{Without support} \\
\noalign{\hrule height 1pt}

Force-balance feasible (\%)
& \textbf{78.9}
& 17.9 \\

Median predicted resistance (N)
& \textbf{13.3}
& 0.0 \\

\noalign{\hrule height 1pt}
\hline
\end{tabular}}
\vspace{-7mm}
\label{tab:floor_allocation}
\end{table}

\begin{table}[t]
  \vspace{2mm}

\caption{Objective ablation for region-constrained geometric refinement on the RH56 under a shared evaluation protocol.}
  \vspace{-2mm}

\centering
\renewcommand{\arraystretch}{1.18}
\scriptsize
\setlength{\tabcolsep}{3.0pt}
\resizebox{0.9\linewidth}{!}
{%
\begin{tabular}{l|cccc}
\hline
\noalign{\hrule height 1pt}
\textbf{Objective}
& \textbf{GSR$\uparrow$}
& \textbf{OSR$\uparrow$}
& \textbf{$F_{\min}\uparrow$}
& \textbf{R@5/10/20/30$\uparrow$} \\
& (\%) & (\%) & (N) & (\%) \\
\noalign{\hrule height 1pt}

A: Geometry only
& \textbf{77.7}
& \textbf{88.5}
& \textbf{10.9}
& \textbf{68.1} / \textbf{53.9} / 37.2 / 27.5 \\

B: Calibrated SOCP
& 66.8
& 76.5
& 8.1
& 58.7 / 45.3 / 29.8 / 22.9 \\

C: Diff. task-GWS ($\pm12$\,mm)
& 73.2
& 80.5
& 10.8
& 61.4 / 52.8 / \textbf{37.5} / \textbf{28.9} \\

C2: Diff. task-GWS ($\pm6$\,mm)
& 72.1
& 85.0
& 10.6
& 60.7 / 51.3 / 33.9 / 25.4 \\

\noalign{\hrule height 1pt}
\hline
\end{tabular}}
\vspace{-3mm}
\label{tab:gws_ablation}
\end{table}







\paragraph{Support contribution analysis}
To quantify how the tabletop contributes to grasp stability, we analyze the settled contact configuration of each grasp using a static second-order cone program (SOCP)~\cite{boyd2007fast}. 
The optimization determines the contact forces at the hand-object and table-object contacts required to balance an external disturbance under friction and motor-torque constraints. 
For each grasp experiment, we solve the same optimization twice: once using both the hand and tabletop contacts, and once after removing only the tabletop contact while keeping the settled hand-object configuration unchanged. 
As shown in Tab.~\ref{tab:floor_allocation}, removing the tabletop substantially reduces force-balance feasibility and disturbance resistance, indicating that the support directly contributes to balancing external forces.

\paragraph{Ablation study on the refinement objective}
We ablate the objective used in the region-constrained geometric refinement in Sec.~\ref{sec:geometric_refinement} by replacing our geometric objective with calibrated SOCP and the differentiable task-GWS objective of 
TaskDexGrasp~\cite{chen2024taskdexgrasp}, which optimizes the grasp wrench space to cover task-specific external wrench directions. The calibrated SOCP is implemented to maximize the worst-case disturbance resistance
certified by the contact-force SOCP~\cite{boyd2007fast} on the pad contacts projected onto the object surface.
Tab.~\ref{tab:gws_ablation} shows that our geometric objective typically yields better physical grasp performance than both analytic alternatives. Although SOCP and differentiable GWS explicitly optimize wrench-based grasp quality, these static objectives can be disconnected from the contacts actually formed after closing and settling.

\begin{table}[t]
\caption{Direction-conditioned resistance after local derivation. Resistance is evaluated along each horizontal direction.}
  \vspace{-2mm}

\centering
\renewcommand{\arraystretch}{1.14}
\scriptsize
\setlength{\tabcolsep}{1.5pt}
\resizebox{0.9\linewidth}{!}
{%
\begin{tabular}{l|ccccccc}
\hline
\noalign{\hrule height 1pt}
\textbf{Condition}
& \textbf{Median}
& \textbf{R@5}
& \textbf{R@10}
& \textbf{R@15}
& \textbf{R@20}
& \textbf{R@25}
& \textbf{R@30} \\
& (N) & (\%) & (\%) & (\%) & (\%) & (\%) & (\%) \\
\noalign{\hrule height 1pt}

Base
& 10.7 & 67.2 & 52.3 & 41.4 & 35.3 & 29.9 & 25.5 \\

$+x$ conditioned
& 16.9 & \textbf{73.5} & \textbf{63.0} & 51.7 & 45.8
& \textbf{41.2} & \textbf{34.0} \\

$-x$ conditioned
& \textbf{17.9} & 73.1 & 61.3 & \textbf{52.9} & \textbf{46.6}
& 38.2 & 33.2 \\

$+y$ conditioned
& 14.4 & 72.7 & 57.1 & 49.2 & 43.3 & 37.0 & 29.4 \\

$-y$ conditioned
& 14.6 & \textbf{73.5} & 60.9 & 49.2 & 42.9 & 38.2 & 32.8 \\

\noalign{\hrule height 1pt}
\hline
\end{tabular}}
\vspace{-7mm}
\label{tab:directional_ascent}
\end{table}

\paragraph{Local derivation for directional robustness} 
We evaluate whether the optional local derivation in Sec.~\ref{sec:derivation} can adapt a validated grasp to a specified disturbance direction. 
We consider four horizontal directions and perform the local search independently for each target direction. 
As shown in Tab.~\ref{tab:directional_ascent}, the derived direction-conditioned grasps exhibit consistently stronger resistance than the original grasps across each tested force. 
This result shows that an already validated grasp can be locally adapted to a task-specific load without repeating the global contact search.

\begin{table}[t]
\caption{Effect of squeeze strength on resistance and settled penetration for fixed RH56 grasps.}
  \vspace{-2mm}
\centering
\renewcommand{\arraystretch}{1.14}
\scriptsize
\setlength{\tabcolsep}{4.0pt}
\resizebox{0.92\linewidth}{!}
{%
\begin{tabular}{l|ccccc}
\hline
\noalign{\hrule height 1pt}
\textbf{Squeeze coefficient}
& \textbf{0.1}
& \textbf{0.3}
& \textbf{0.5}
& \textbf{0.8}
& \textbf{1.2} \\
\noalign{\hrule height 1pt}

$F_{\min}$ median (N)
& 8.0 & 10.0 & 12.4 & 14.1 & \textbf{16.5} \\

6-dir. success (\%)
& 86 & 88 & \textbf{89} & \textbf{89} & 86 \\

Settled penetration median (mm)
& 0.09 & 0.09 & 0.09 & 0.09 & 0.09 \\

\noalign{\hrule height 1pt}
\hline
\end{tabular}}
\vspace{-2mm}
\label{tab:squeeze_intervention}
\end{table}

\paragraph{Effect of squeeze strength} 
We analyze the effect of closing strength by keeping the grasp pose, object, and evaluation protocol fixed while varying only the squeeze coefficient. 
As shown in Tab.~\ref{tab:squeeze_intervention}, stronger squeezing consistently increases weakest-direction resistance, while grasp success and settled penetration remain nearly unchanged. 
In particular, the median penetration remains constant across the entire squeeze range. 
This shows that the increased resistance is primarily associated with stronger contact loading rather than deeper penetration into the object.

\begin{figure*}[t]
  \centering
  \vspace{2mm}  \includegraphics[width=2\columnwidth, clip]{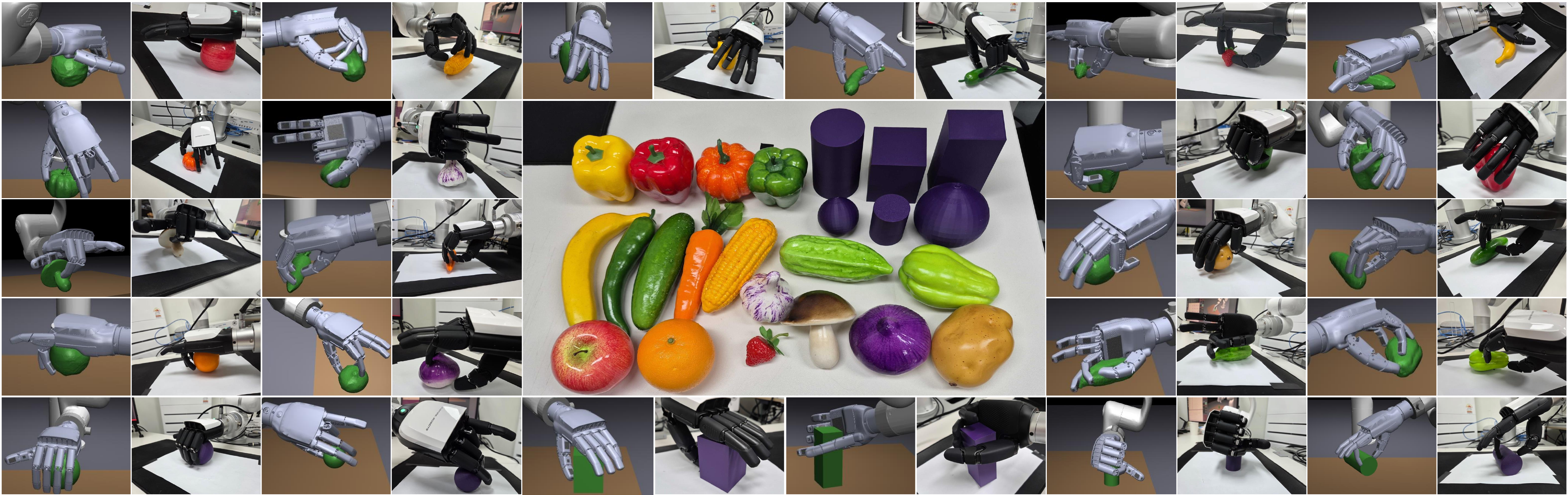}
   \vspace{-3mm}
    \caption{Qualitative results of grasping real objects. The center image displays real objects used in the experiments and the surrounding images illustrate diverse grasp configurations adapting to each object shape.}
    \label{fig:grasp_rigid}
   \vspace{-5mm}
\end{figure*}

\begin{figure}[t]
  \centering
  \includegraphics[width=0.98\columnwidth, clip]{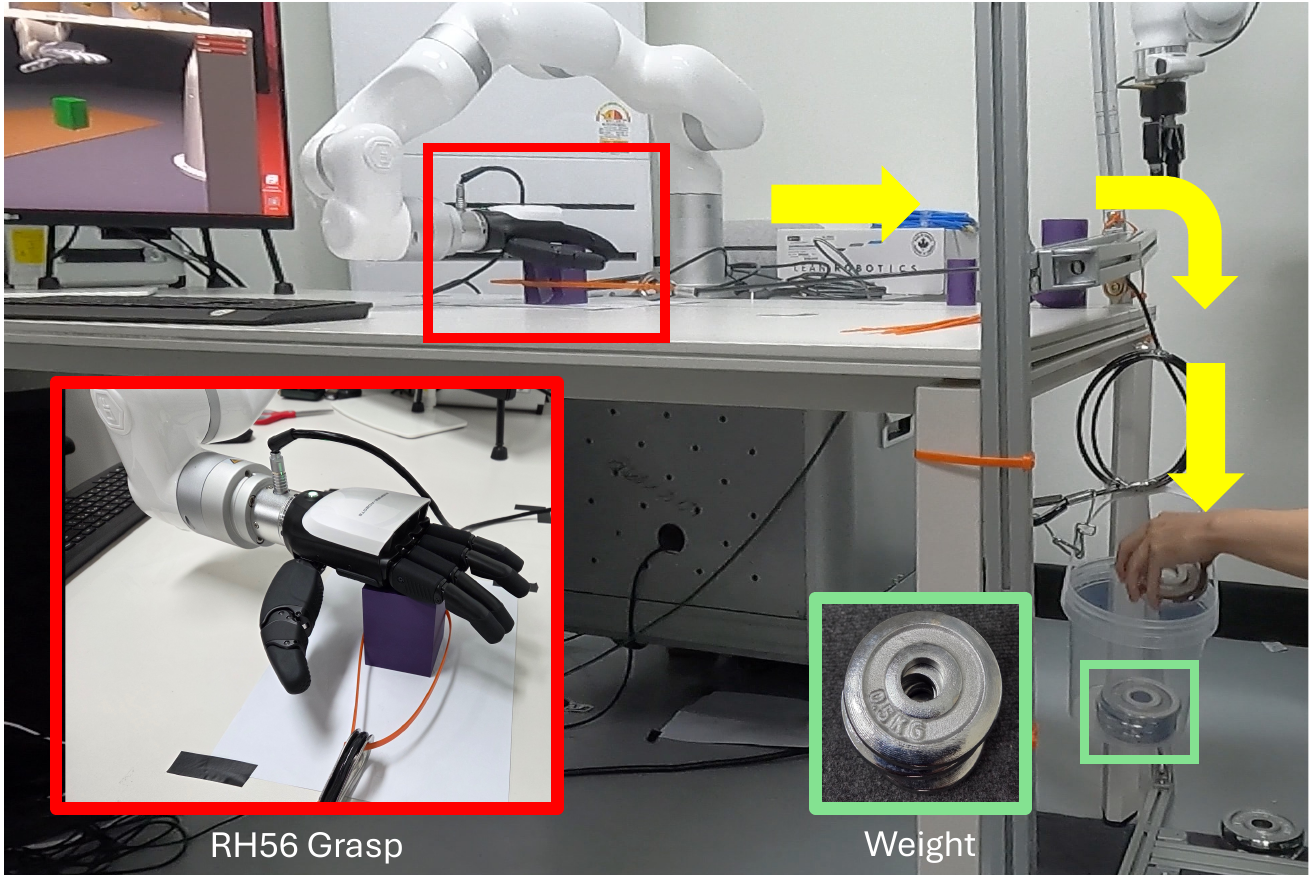}
  \vspace{-3mm}
    \caption{Experimental setup for evaluating disturbance resistance in the real world. The yellow arrow denotes the disturbance force. This force is caused by the gravity of the weights and transmitted to the objects via a pulley.}
    \label{fig:exp_weight}
   \vspace{-7.5mm}
\end{figure}

\vspace{-1mm}
\subsection{Real-World Evaluation}
\label{sec:real_world}

All real-world experiments are shown in the video.

\paragraph{Real-world grasp execution}
We first evaluate whether the synthesized grasps are executable on the real RH56 robot across diverse object geometries. 
We test 24 fruit and vegetable replicas using grasps that span fingertip, proximal-link, palm, and table-assisted contacts.
Fig.~\ref{fig:grasp_rigid} shows representative simulated configurations together with their physical executions. 
The resulting grasps preserve diverse contact configurations on the real hand for various object shapes, demonstrating that the proposed synthesis is not restricted to a simple grasp pattern and can produce physically executable configurations for the coupled RH56.

\begin{table}[t]
\vspace{-1mm}
\caption{Real-world evaluation of grasp success and disturbance resistance. $R@F$ denotes the success rate under forces $F$.}
\vspace{-2mm}
\centering
\renewcommand{\arraystretch}{1.12}
\scriptsize
\begin{tabular*}{\linewidth}{@{\extracolsep{\fill}}c|cccccc@{}}
\hline
\noalign{\hrule height 1pt}
\textbf{Object}
& \textbf{GSR}
& \textbf{R@10}
& \textbf{R@15}
& \textbf{R@20}
& \textbf{R@25}
& \textbf{R@30} \\
\noalign{\hrule height 1pt}

1 & 100  & 100  & 100  & 66.7 & 33.3 & 33.3 \\
2 & 100  & 100  & 100  & 66.7 & 33.3 & 33.3 \\
3 & 66.7 & 66.7 & 0.0  & 0.0  & 0.0  & 0.0 \\
4 & 100  & 100  & 66.7 & 66.7 & 66.7 & 66.7 \\
5 & 100  & 100  & 100  & 66.7 & 66.7 & 66.7 \\
6 & 100  & 100  & 100  & 100  & 100  & 100  \\

\noalign{\hrule height 1pt}
\textbf{Overall}
& \textbf{94.4}
& \textbf{94.4}
& \textbf{77.8}
& \textbf{61.1}
& \textbf{50.0}
& \textbf{50.0} \\

\noalign{\hrule height 1pt}
\hline
\end{tabular*}

\vspace{-3mm}
\label{tab:real_force}
\end{table}

\paragraph{Real-world disturbance resistance} 
We evaluate six objects with three different synthesized grasps per object, resulting in 18 physical grasp configurations. As demonstrated in Fig.~\ref{fig:exp_weight}, each object is connected to a hanging weight through a pulley, and the disturbance is progressively increased by adding weights up to $30$\,N. 
We measure both grasp success and the stability of grasps that remain settled under increasing external loads. 
As shown in Tab.~\ref{tab:real_force}, 17 of the 18 grasps are successfully executed, and the resistance remains high across multiple grasp configurations and objects. More than half of all grasps resist $20$\,N, while half remain stable at the maximum tested load of $30$\,N. These results demonstrate that the synthesized grasps provide substantial physical resistance across objects and grasp configurations.

\begin{figure}[t]
  \centering
  \includegraphics[width=.98\columnwidth, clip]{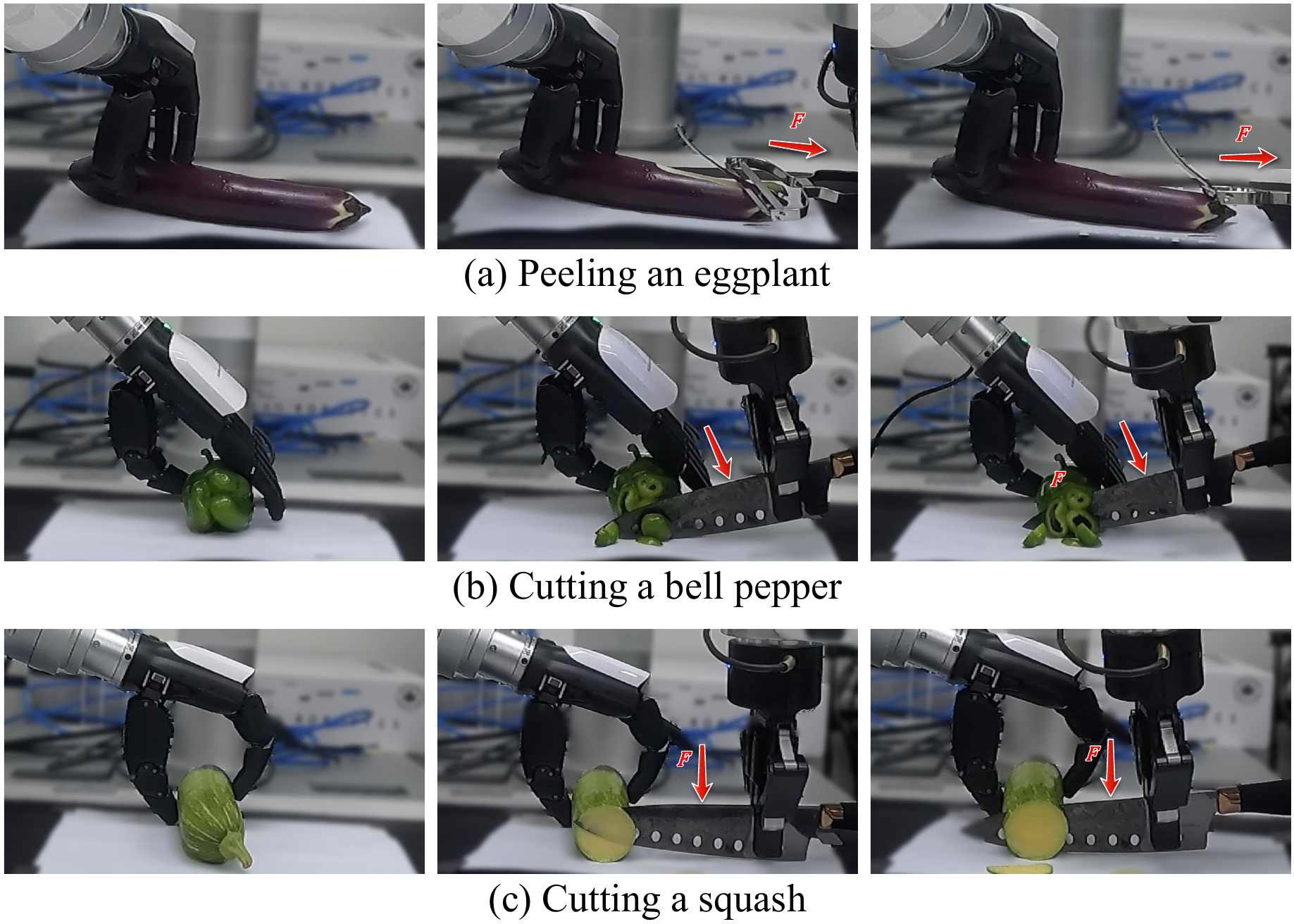}
\vspace{-3mm}
    \caption{Demonstration of real-world deployment in a cooking scenario. Our synthesized grasps demonstrate high stability against the forces caused by cutting or peeling vegetables. }
    \label{fig:real_cutting}
\vspace{-7.5mm}

\end{figure}

\paragraph{Downstream manipulation}\
Finally, we demonstrate a downstream manipulation application using six real vegetables. Before execution, we apply the optional task-guided local derivation (Sec.~\ref{sec:derivation}) to obtain a support-assisted grasp that avoids the region required for tool access. As shown in Fig.~\ref{fig:real_cutting}, the RH56 holds the object against the tabletop while a knife or slicer is teleoperated to perform cutting~\cite{wu2024gello}. 
This experiment illustrates that the proposed local derivation can adapt a validated grasp to task-specific constraints and demonstrates that the derived grasps remain stable enough to support subsequent physical interactions such as cutting vegetables.
Moreover, as cutting changes the object geometry, the same framework can be applied to the updated shape to derive a new grasp configuration for continued manipulation.

\vspace{-1.0mm}

\section{CONCLUSIONS}
We present a support-assisted grasp synthesis framework for low-DoF coupled robotic hands. 
Our key idea is to search for contact structures that are compatible with the coupled hand kinematics, refine them toward geometrically realizable contacts, and validate the resulting grasps through physical execution. 
Rather than treating the surface only as a geometric constraint, our framework allows it to provide additional support when forming stable grasps. 
Experiments on the RH56 and Shadow Hand show that accounting for hand embodiment and environmental support improves grasp feasibility and robustness, while real-robot results show that the synthesized grasps remain executable and adaptable to downstream tasks.
Our current pipeline focuses on static tabletop support and known object geometry. 
Extending the pipeline to more general environmental contacts along with perception-driven object and support geometries would broaden its applicability to less structured manipulation scenarios.
In addition, the synthesized grasps could be organized into a large-scale dataset or used as structured grasp supervision to inject grasp knowledge into robot learning systems.



\vspace{-1mm}

\bibliographystyle{IEEEtran}
\bibliography{refs}
\end{document}